\documentclass[final]{cvpr}
\usepackage{textcomp}
\usepackage{gensymb}
\usepackage{times}
\usepackage{epsfig}
\usepackage{graphicx}
\usepackage{amsmath}
\usepackage{amssymb}
\usepackage{xparse}
\usepackage[thinlines]{easytable}
\usepackage{xcolor}
\usepackage{amsmath}
\usepackage{balance}

\newcommand{\nnl}{s}
\newcommand{\fell}{f^{\nnl}}
\newcommand{\fellm}{f^{\nnl-1}}
\newcommand{\chiell}{h^{\nnl}}

\usepackage{mathtools}

\usepackage[pagebackref=true,breaklinks=true,colorlinks,bookmarks=false]{hyperref}

\def\cvprPaperID{9} 
\def\confYear{CVPR 2021}

\usepackage{cuted}
\usepackage{capt-of}

\begin{document}
\title{OmniLayout: Room Layout Reconstruction from Indoor Spherical Panoramas}

\author{Shivansh Rao \thanks{equal contribution} \qquad Vikas Kumar \footnotemark[1] \qquad Daniel Kifer\qquad C. Lee Giles\qquad Ankur Mali\\The Pennsylvania State University, University Park\\
PA, USA 16802\\
\{\tt\small shivanshrao,vuk160,duk17,clg20,aam35\}@psu.edu
}

\maketitle

\begin{strip}\centering
\includegraphics[width=\textwidth]{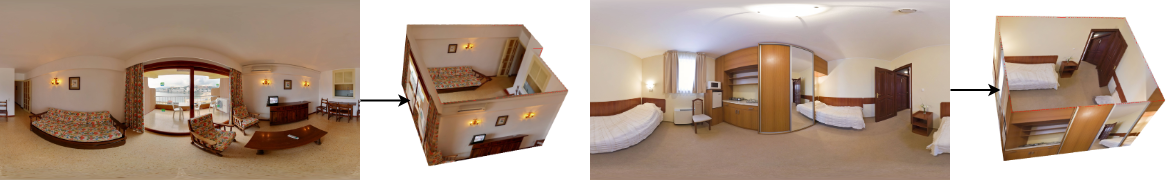}
\captionof{figure}{\textbf{Illustration: }Our OmniLayout can predict both non-cuboid layout as well as cuboid layout from the input RGB panorama.
\label{fig:feature-graphic}}
\end{strip}

\begin{abstract}
 
Given a single RGB panorama, the goal of 3D layout reconstruction is to estimate the room layout by predicting the corners, floor boundary, and ceiling boundary. A common approach has been to use standard convolutional networks to predict the corners and boundaries, followed by post-processing to generate the 3D layout. However, the space-varying distortions in panoramic images are not compatible with the translational equivariance property of standard convolutions, thus degrading performance. Instead, we propose to use spherical convolutions. The resulting network, which we call OmniLayout performs convolutions directly on the sphere surface, sampling according to inverse equirectangular projection and hence invariant to equirectangular distortions. Using a new evaluation metric, we show that our network reduces the error in the heavily distorted regions (near the poles) by $\approx 25 \%$ when compared to standard convolutional networks. Experimental results show that OmniLayout outperforms the state-of-the-art by $\approx$4$\%$ on two different benchmark datasets (PanoContext and Stanford 2D-3D). Code is available at \textcolor{red}{ https://github.com/rshivansh/OmniLayout}.
\end{abstract}


\section{Introduction}\label{sec:intro}
Estimating the 3-dimensional layout of the room from a single RGB image has received considerable attention in the last decade. Layout estimation can present useful information (height, corner positions, and orientation of the room)  for holistic scene understanding applications such as robotics and augmented/virtual reality \cite{sun2020hohonet,fernandez2020indoor}. Most of the previous works \cite{sun2019horizonnet,yang2019dula,zou2018layoutnet} tackle room layout estimation problem by using artificial neural networks (ANNs).  They capture the salient features from the image while considering the manhattan room layout \cite{coughlan1999manhattan}. These approaches have shown impressive results not just in terms of the quantitative evaluation but also qualitatively by generating both cuboid-shaped room layouts as well as non-cuboid-shaped general layouts. Since conventional cameras have a limited field of view leading to several ambiguities, existing literature \cite{sun2019horizonnet,yang2019dula,zou2018layoutnet, fernandez2020corners} directly operates on 360$\degree$ panoramas, exploiting the wider field of view.


Although existing work are heavily dependent on standard convolution layers, showing impressive results on few panoramic benchmarks \cite{sun2019horizonnet,yang2019dula,zou2018layoutnet}. We believe standard convolution often fail to capture features in panoramic images, thus leading to sub-optimal representation. Prior works also argue that standard convolutions are not well suited for processing panoramic images \cite{coors2018spherenet,eder2019mapped}. This is because equirectangular images, which are considered a common example for spherical image representation, have heavy distortions in them (especially towards the poles) which cannot be addressed by standard convolutions \cite{coors2018spherenet, eder2019mapped}. Thus leading to bottleneck in all the prior approaches while dealing with room layout estimation. 
To address this problem and towards building robust representation for panoramic image we present OmniLayout: a deep neural network that estimates the room layout while accounting for these common distortions pattern. Inspired from \cite{coors2018spherenet}, our model tackles the distortions in the given equirectangular image by changing where the convolutional kernel samples from the image in a location-dependent manner. Instead of performing the convolution operation on the regular image domain, our network performs convolutions on the sphere surface where the omnidirectional images can be represented without any distortions. We show that our methodology significantly boosts the performance in the complex regions of the images (i.e., the polar regions containing most of the distortions) while maintaining equal or better performance in the least complex regions of the images (near the equator).

While Coors \textit{et al.} \cite{coors2018spherenet} uses gnomic projection to map the sphere onto a tangent plane, we argue that this is not an accurate projection for equirectangular images. Instead, we perform sampling using inverse equirectangular projection which leads to better representation across wide variety of networks, as shown in later section. Our main idea is to use more principled approach and replace the standard convolutions with spherical convolutions, which we believe are well-suited for the task of room layout estimation from a panorama. We build our network on top of HorizonNet \cite{sun2019horizonnet} and replace standard convolution operation with spherical convolution for enhanced representation and reduce computational complexity by replacing Bi-LSTM with Bi-GRU. We validate our hypothesis by conducting several experiments across two large-scale benchmarks \cite{zhang2014panocontext, armeni2017joint}. Finally, we conduct an ablation study across each model component to highlight their significance and contribution resulting in better estimation over panoramic images.

\section{Related Work}\label{sec:related work}

Room layout estimation from a single RGB image has been an active area of research in the last decade. The existing literature differs in mainly two different aspects: 1) input image type, and 2) proposed methodology. In this section, we review several lines of
related work falling in each of the categories.


In terms of input image type, prior work differ on the basis of the field of view (FoV), ranging from the normal perspective images to 360$\degree$ panorama images. Delage \textit{et al.} \cite{delage2006dynamic},  Hedau \textit{et al.} \cite{hedau2009recovering} and Lee \textit{et al.} \cite{lee2009geometric} operate only on the perspective images, while Zhang \textit{et al.} \cite{zhang2014panocontext} estimates the room layout directly from a single 360$\degree$ panorama and proposes the PanoContext dataset. Xu \textit{et al.} \cite{xu2017pano2cad} combines surface normal estimates, 2D object detection, and 3D object pose estimation to estimate the room layout and 3D pose of the object. There are some other works that use more information than just a single image, such as using multiple images \cite{cabral2014piecewise} or using the depth information as well (RGB-D data) \cite{liu2016layered, guo2015predicting, zhang2013estimating}.

Most of the recently proposed methodologies incline towards adopting deep neural networks to improve layout estimation. These approaches use dense models to predict the semantic label of each pixel. Some of these approaches \cite{mallya2015learning, ren2016coarse, zhao2017physics} operate on the perspective images. Mallya \textit{et al.} \cite{mallya2015learning} learns to predict informative edge probability maps whereas Zhao \textit{et al.} and Ren \textit{et al.} \cite{zhao2017physics, ren2016coarse} predict for the boundary classes. Since the recent increase in omnidirectional sensors, there have been a few deep learning approaches that directly operate on panoramas. Zou \textit{et al.} \cite{zou2018layoutnet} presents a method that can generate both cuboid layout and general layout directly from the given panorama. Yang \textit{et al.} \cite{yang2019dula} uses two different projections of the panorama at the same time (front-view panorama and top-view perspective) showing the advantages of additional information from the ceiling-view image. Sun \textit{et al.} \cite{sun2019horizonnet} presents a new approach by representing the room layout as a 1D representation.

Although existing works show impressive performance for both cuboid as well as non-cuboid layouts \cite{sun2019horizonnet, yang2019dula, zou2018layoutnet}, none of them considers the distortions that the equirectangular images contain. There is an incongruence between the panoramic images and standard convolutional networks. A few recent approaches have proposed to overcome the distortions by using spherical convolutions. Su \textit{et al.} \cite{su2017learning} proposes to increase the kernel size of the standard convolution filters towards the polar regions. However, this results in a significant increase in the model parameters, since the weights now can only be shared along each row. Cohen \textit{et al.} \cite{cohen2018spherical} proposes to use spherical CNNs that encodes full rotational invariance. However, assuming that the camera is not tilted while capturing $360\degree$ images, full rotational invariance is an undesired property for our task and reduces the discriminative power of the model. In concurrent work, Coors \textit{et al.} \cite{coors2018spherenet} addresses the issue by capturing rotational invariance only in one dominant orientation and is compatible with modern CNN architectures. Additionally, it allows the transfer of pre-trained object detectors to omnidirectional inputs. Results show that SphereNet \cite{coors2018spherenet} performs better than the other methods that handle omnidirectional inputs on benchmark datasets Omni-MNIST and FlyingCars \cite{coors2018spherenet}.

\begin{figure*}[!t]
    \centering
    \includegraphics[width=\textwidth]{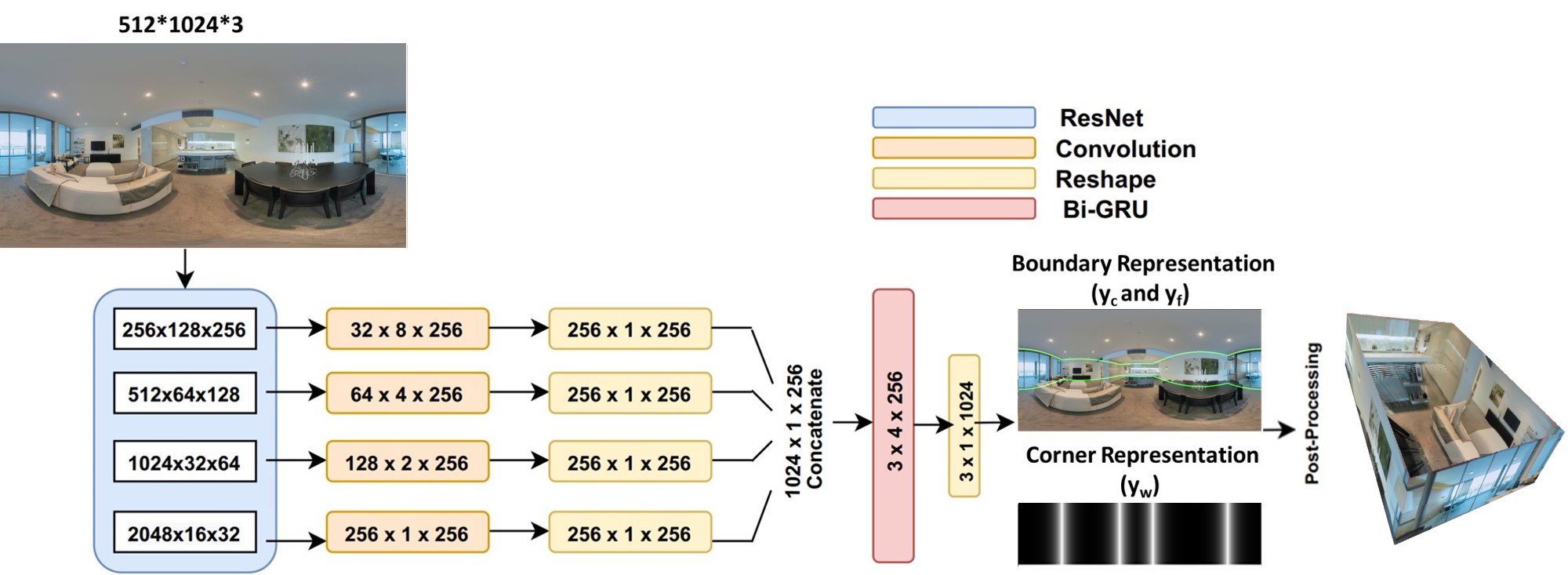}
    \caption{\textbf{OmniLayout architecture:} Our model is built on ResNet followed by Bi-Directional GRU that predicts the positions of corners of the room. We have replaced the standard convolution from each block of ResNet with the sphere convolution (inverse equirectangular projection). The output of the network is a 1D representation map of shape $3 \times 1 \times 1024 $. Since the width of the panorama is 1024 the output map has 3 values per column: $y_c$ (ceiling-wall), $y_f$ (floor-wall) and $y_w$ (wall-wall / corners).}
    \label{fig:arch}
\end{figure*}

To our knowledge, none of the previous work in room layout estimation deal with the shortcoming that standard convolutions have in terms of panoramic images, except the work in \cite{fernandez2020corners}. While Clara \textit{et al.} \cite{fernandez2020corners} has attempted to reduce the distortions in the equirectangular representation, it considers the convolutional kernel as a tangent plane to a sphere making use of the inverse gnomic projection which is not the accurate projection type for equirectangular images. In our work, we use the inverse equirectangular projection which we believe is the accurate projection type to eliminate the existing distortions in equirectangular images.

\section{Approach}\label{sec:app}

In this section, we describe our end-to-end network for generating the 3D room layout from a single RGB panorama. We first give a brief overview of our architecture (Sec. \ref{subsec:brief}), followed by the description of inverse equirectangular projection for spherical convolutions (Sec. \ref{subsec:sphereconv}). Then we describe our model's architecture (Sec. \ref{subsec:enc}, \ref{subsec:rnn}). Finally, we describe the post-processing details for generating the 3D room layout from the model's predictions (Sec. \ref{subsec:3D}).

\subsection{Network Architecture}\label{subsec:brief}
An overview of OmniLayout is illustrated in Fig. \ref{fig:arch}. The proposed architecture consists of a ResNet-50 \cite{he2016deep} encoder with proposed spherical convolutions. We remove the final fully-connected layer and concatenate the features from different levels and pass it to a Bi-Directional Gated Recurrent Unit (Bi-GRU) \cite{chung2014empirical} that predicts the layout floor-wall boundary ($y_f$), ceiling-wall boundary ($y_c$), and wall-wall boundary ($y_w$).

\subsection{Convolution for Panoramic Images}
\label{subsec:sphereconv}
\begin{figure}[!t]
    \centering
    \includegraphics[width=\linewidth]{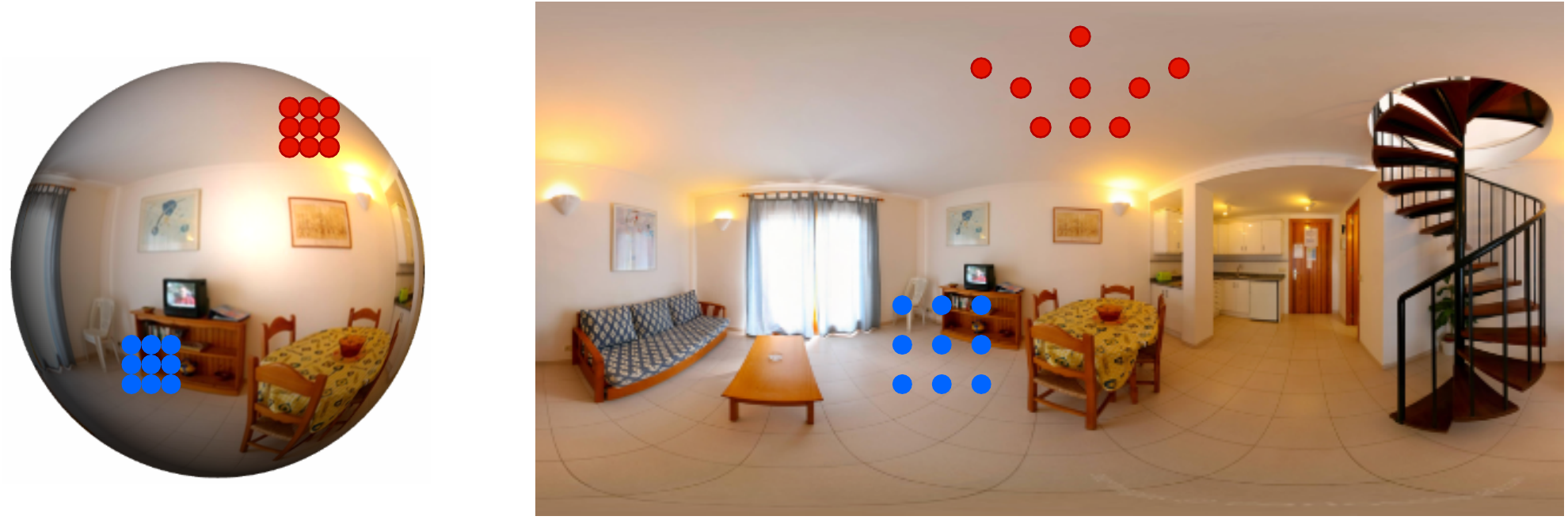}
    \caption{\textbf{Spherical Convolution on panoramic images:} We show two different kernel positions, one at the center of the image (\textcolor{blue}{blue}) and one toward the poles (\textcolor{red}{red}). Equirectangular images usually has more distortions near the poles. The proposed kernel deforms itself near the poles accounting for the distortions in that region when compared to the region near the equator.
}
    \label{fig:equiconv}
\end{figure}

Omnidirectional sensors have gained huge popularity in the last few years due to their wider field of view with several applications in virtual/augmented reality and robotics. Due to an increase in omnidirectional sensors, spherical imagery is receiving increased attention as well. The most common representation of spherical images is the equirectangular projection in which the longitude and latitude of a spherical image are mapped to vertical and horizontal coordinates. However, this mapping comes with heavy distortions, especially near the poles. Standard convolutions are not a good choice for such images. From Fig. \ref{fig:equiconv} we can observe how the proposed kernel deforms itself near the poles in order to account for the distortions.



One of the simplest examples of a covariant neural network one can consider are traditional $s+1$ layers CNN used for image recognition and other vision-related tasks.  
Traditionally neurons in each layer of CNN are arranged in a rectangular grid. Let us consider a network with a single channel, then the activation of layer ${s}$ can be regarded as a function 
${\fell\colon \mathbb{Z}^2 \to \mathbb{R}}$, with ${f^0}$ being the input image \cite{cohen2018spherical,kondor2018clebschgordan}. We now adopt notations and definitions proposed in prior work \cite{kondor2018clebschgordan} and define the overall flow for spherical CNNs. 
As noted earlier the neurons in our network compute ${\fell}$ by taking the cross-correlation
of the previous hidden layer's output ${\fellm}$ with a learnable filter or kernel ${\chiell}$ as follows:

\begin{equation}\label{eq: xcorr1}
(\chiell \star \fellm)(x)=
\sum_{y} \chiell(y - x)\;\fellm(y)
\end{equation}

then we apply nonlinear activation function $\sigma$, such as the ReLU or other variants operator \footnote{Better results can be obtained by using variants of ReLU or other functions such as SELU which might lead to better gradient flow across model.}:

\begin{equation}\label{eq: CNN1}
\fell(x)=\sigma(\chiell \star \fellm)(x)
\end{equation}\newline

Defining: $T_x(\chiell)(y)=\chiell(y-x)$, which is  
nothing but ${\chiell}$ translated by ${x}$, allows us to equivalently write  Eq.\ref{eq: xcorr1} as follows:
\begin{equation}\label{eq: xcorr2}
(\chiell \star \fellm)(x)=\fellm,T_x(\chiell), 
\end{equation}

where the inner product is $\fellm$,  $T_x(\chiell)=\sum_{y} \fellm(y)\,T_x(\chiell)(y).$
This formulation as noted in prior works indicates that each layer in CNN are doing some kind of pattern matching: 
${\fell(x)}$ is an indicatior of how well the part of ${\fellm}$ around ${x}$ matches the filter or kernel ${\chiell}$. Equation \ref{eq: xcorr2} is the natural starting point for generalizing
convolution to the unit sphere, ${S^2}$.

A number of authors have addressed the issue of discretizing ${S^2}$ by a regular arrangement of points, which is often convenient when dealing with planes  \cite{boomsma2017spherical,su2017learning}. 
Instead of following the aforementioned approaches, similarly to recent work on manifold CNNs 
\cite{masci2015geodesic,monti2017geometric}, 
one can simply treat each ${\fell}$ and the corresponding filter ${\chiell}$ 
as continuous functions on the sphere \cite{kondor2018clebschgordan}, ${\fell(\theta,\phi)}$ and ${\chiell(\theta,\phi)}$, 
where ${\theta}$ and ${\phi}$ are the polar and azimuthal angles. 
Thus, we perform the convolutions operations directly on the sphere surface instead of the image domain, giving use advantage when compared with competitors.
We allow both functions to be complex-valued which is argued to provide better generalization \cite{kondor2018clebschgordan}.

Finally the correct way to generalize cross-correlations on a sphere while considering the rotation around a third axis \cite{kondor2018clebschgordan} can be established by defining ${h \star f}$ as a function, that is represented as follows:

\begin{equation}\label{eq: S^2 xcorr}
(h \star f)(R)=
4\pi\int_{0}^{2\pi}\int_{-\pi}^{\pi} h_R(\theta,\phi)^\ast
f(\theta,\phi)\cos\theta d\theta d\phi
R(3),
\end{equation}

where ${h_R}$ is ${h}$ rotated by ${R}$. Further we can express these terms as follows:
\begin{equation}\label{eq: rotation}
h_R(x)=h(R^{-1}x),
\end{equation}
with ${x}$ being the point on the sphere at position ${(\theta,\phi)}$ . 

\begin{figure}
    \centering
    \includegraphics[height=4cm,width=0.8\linewidth]{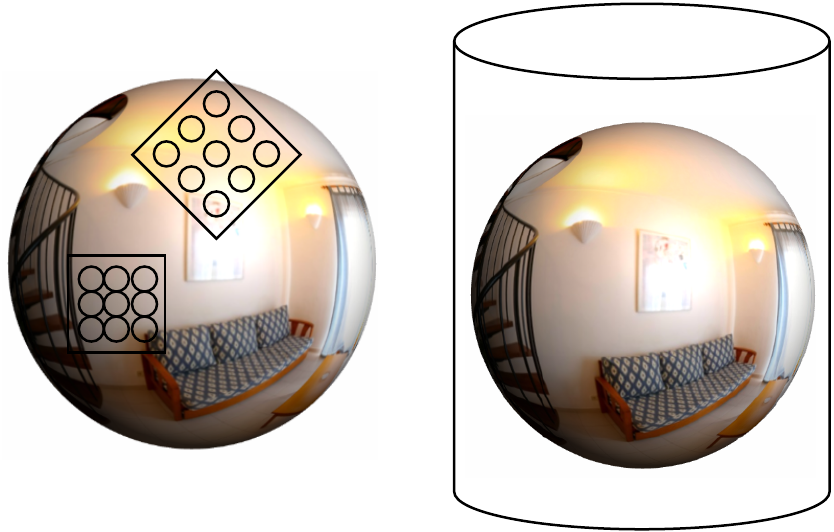}
    \caption{Gnomic projections are azimuthal projection (left) that project sphere to tangent planes, and Equirectangular projections are cylindrical projection (right) that project sphere to a cylinder.}
    \label{fig:projection}
\end{figure}

This formulation offers one key advantage by efficiently encoding equirectangular projection into the kernel's sampling function, thus allowing better estimation over the spherical surface as opposed to standard convolutions. We formulate the kernel over a cylindrical patch available in the spherical surface and then sample the equirectangular projection. The positions of the kernel locations on the cylindrical patch are calculated similarly to \cite{coors2018spherenet}, while ensuring that we use equirectangular projection instead of gnomic projection. The equirectangular projection is described as follows:

\small
\begin{equation}\label{eq:equiconv}
  \begin{aligned}
  \theta = v_{0} + \Delta v_{(i,j)},\\
  \phi = u_{0} + \Delta u_{(i,j)}\sec \theta,\\
  \end{aligned}
\end{equation}

where the sphere is parameterized in terms of its polar ($\theta$) and azimuthal angles ($\phi$). u$_{0}$ and v$_{0}$ represents the center of the kernel, $\Delta$ u$_{(i,j)}$ and $\Delta v_{(i,j)}$ represents the angular distance at index (i,j) from the kernel center in the x and y direction respectively. The approach proposed by Clara \textit{et al.} \cite{fernandez2020corners} and Coors \textit{et al.} \cite{coors2018spherenet} instead utilizes the inverse gnomic projection which maps the sphere to a tangent plane (See Fig. \ref{fig:projection}).
Since equirectangular images are cylindrical projections that project sphere to a cylinder, the distortions produced by them are different which can not be handled by the gnomic projections.



%

\subsection{Encoder}\label{subsec:enc}
To be comparable with current state-of-the-art model, we adopt the same feature extractor - ResNet \cite{he2016deep} as the HorizonNet \cite{sun2019horizonnet}. The input panorama is of shape \textit{$3\times512\times1024$}. ResNet initially has a convolution layer of $7 \times 7$ kernel with stride 2 and padding 3. This is followed by four blocks, each block consists of a sequence of convolution layers reducing the channels and height by a factor of 8 (i.e. for first block 256 / 8 = 32) and 16 (i.e. for first block 128 / 16 = 8) respectively. Precisely, there are three convolution layers in each block. The features from different blocks help to capture both low-level, as well as high-level information \cite{kumar2020noisy} from the given panorama. The output feature from each block is reshaped to \textit{$256 \times 1 \times 256$} tensors and concatenated to form a single tensor of shape \textit{$1024 \times 1 \times 256$}. In the base architecture of ResNet, we convert all the standard convolutions to spherical convolutions with inverse equirectangular projections (see section \ref{subsec:sphereconv}). In the ablation study (see section \ref{subsec:ablation}), we show that the spherical convolution shows improvement across the entire family of ResNet and is not restricted to ResNet-50.

\subsection{Recurrent Neural Network}\label{subsec:rnn}

\begin{figure}[!t]
    \centering
    \includegraphics{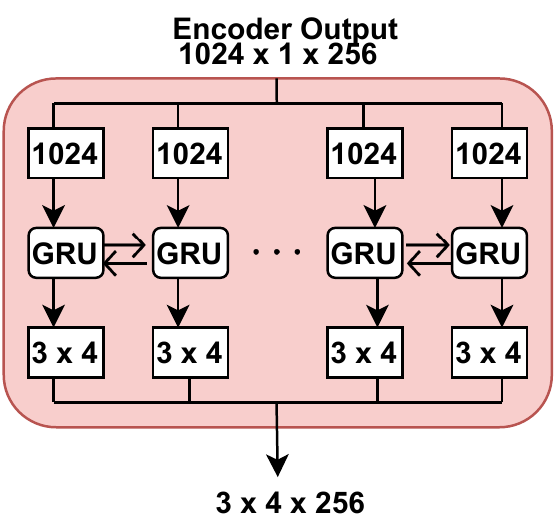}
    \caption{Illustration of Bi-GRU used for predicting corners, floor boundary and ceiling boundary.}
    \label{fig:gru}
\end{figure}

Due to the geometry of a room, a corner can be approximately predicted from the position of other corners of the room. Assuming this we feed the concatenated feature map from encoder as the input sequence to a recurrent neural network (RNN), more specifically to a bi-directional gated recurrent unit (Bi- GRU). RNN's are stateful models better known for capturing long-range dependencies. Non-local neural networks \cite{rao2019non,wang2018non} are another alternative and are faster in comparison to RNN's, however we leave this for future investigation \footnote{We performed experiments with both LSTM \cite{hochreiter1997long} as well as GRU \cite{chung2014empirical} and finalized Bi-GRU for our network since it trains faster and gives approximately the similar performance as a Bi-LSTM.}. The input sequence is of shape $1024 \times 1 \times 256$ and the Bi-GRU produces the output sequence of shape $3 \times 4 \times 256$ which is later reshaped to $3 \times 1 \times 1024$ (see Fig. \ref{fig:arch} and Fig. \ref{fig:gru}). Thus the room layout is represented as three 1D predictions similar to \cite{sun2019horizonnet}. This formulation leads to computational efficiency model while training.

We set Bi-GRU sequence length equal to the width of the image (1024) and predicts three values for each column of the image ($y_c$, $y_f$, and $y_w$). To reduce the computational time the Bi-GRU predicts for four columns at any given time instead of a single column thus the output is of shape $3 \times 4 \times 256$. We use the bidirectional nature of GRU since it offers flexibility of incorporating left and right context values ($y_w$, $y_f$, and $y_c$) offering enhanced representation for our model.

\subsection{3D Layout Generation}\label{subsec:3D}
For recovering the 3D layouts from the predictions, we follow the methodology described by Sun \textit{et al.} \cite{sun2019horizonnet} and make the following assumptions: (a) all rooms follow the Manhattan world assumptions, (b) the camera height is 1.6 meters \cite{zhang2014panocontext} above the floor, and (c) pre-processing \cite{zou2018layoutnet} correctly aligns the floor perpendicular to the y-axis. There are two broad steps in the layout recovery, the first is to recover the floor plane and ceiling plane, while the second is to recover the wall-wall planes. First the model's predictions provide the locations of floor boundaries ($y_{f}$), and ceiling boundaries ($y_{c}$) for every column, we can project them from image coordinates to 3D XYZ coordinates. Second the ceiling wall boundaries share the same positions as the floor wall boundary (X and Z). We then subtract the ceiling and floor 3D coordinates for each image column and take the average over all the image columns to get the height $h$ of the room.

\begin{figure}[!t]
    \centering
    \includegraphics[width=\linewidth]{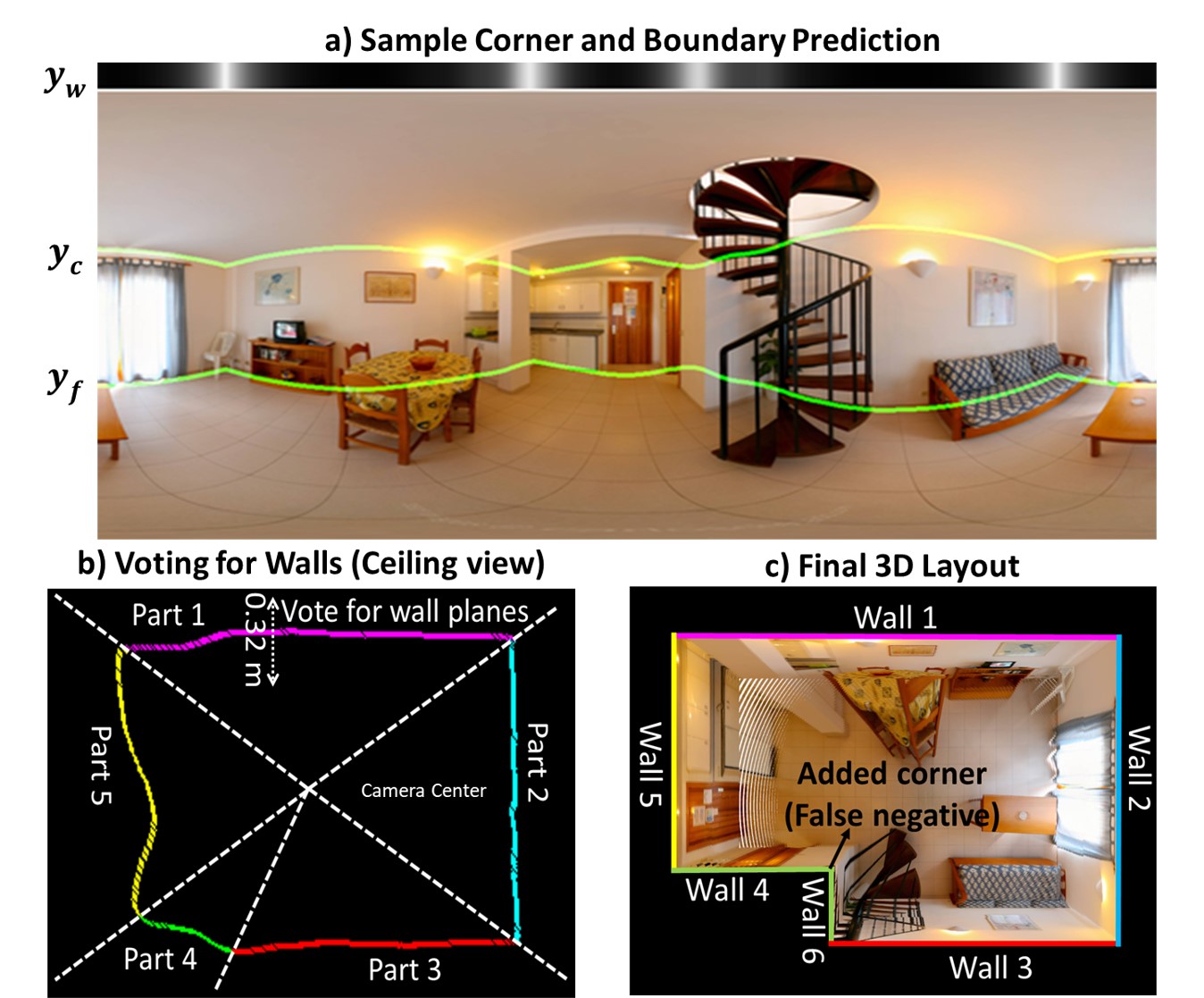}
    \caption{Our model predicts (top) the ceiling-wall ($y_c$) boundary, floor-wall ($y_w$) boundary and the wall-wall ($y_w$) probability map. The post-processing is done in ceiling view (bottom-left). This helps in enforcing the orthogonality of adjacent walls and helps in detecting false negative corners (bottom-right).}
    \label{fig:3d_layout}
\end{figure}

Later the wall planes are recovered by selecting peak points from the predicted wall-wall probability map ($y_w$) which have the peak signal strength in its 5$\degree$ field of view and minimum signal strength of 0.05. While the prediction of boundaries and corners are done using the equirectangular view (See Fig. \ref{fig:3d_layout}a), the post-processing is done using the ceiling view. To correct the horizontal alignment of the 3D layout, the ceiling wall boundary is divided into parts ( $p_1, p_2, ..., p_n$ ) using the prominent peaks (see Fig. \ref{fig:3d_layout}b). It then gives a higher score to the vector line with more pixel points within 0.16 meters and selects the vector that obtains the highest score as the wall for every part $p_i$ (see Fig. \ref{fig:3d_layout}b).

Finally we force adjacent walls to be orthogonal to each other, however the wall whose adjacent walls have not been constructed yet are free to choose the orthogonality type. We also consider special cases, where two adjacent walls for a part $p_i$ are already constructed, but their vector lines are orthogonal to each other (instead of being parallel). This may occur in rare cases of occluded or undetected corners, hence its important to add an additional corner to the layout based on the position of the adjacent walls with respect to part $p_i$. For example, in Figure \ref{fig:3d_layout}, we see that Wall 5 is orthogonal to Wall 3 which violates the Manhattan properties leading to a new corner being added to the layout. The ceiling-wall corner points are established using the intersection of 3 perpendicular planes (2 adjacent walls and a ceiling). The floor-wall corner points are found using the ceiling-wall corner points and the height of the room. More details can be found in \cite{sun2019horizonnet}.

\section{Experiments}\label{sec:exp}

In this section, we first introduce the datasets used in the experiments (Sec. \ref{subsec:data}). Then we describe our experimental setup and some implementation details (Sec. \ref{subsec:training_Details}).
At the end, we present the experimental results and compare with other state-of-the-art results (Sec. \ref{subsec:results} and Sec. \ref{subsec:ablation}).

\subsection{Dataset}\label{subsec:data}
We conduct experiments on two benchmark datasets: PanoContext \cite{zhang2014panocontext} and Stanford 2D-3D \cite{armeni2017joint} extended by \cite{zou2018layoutnet}.

\textbf{PanoContext:} This dataset consists of 500 annotated cuboid room layouts. We perform the same experimental protocol as \cite{sun2019horizonnet} and \cite{zou2018layoutnet} by
splitting 10 $\%$ validation images from the training set to make sure similar rooms do not appear in the training set. The panoramas are captured from indoor settings such as living rooms and bedrooms.

\textbf{Stanford 2D-3D:} This dataset consists of 571 RGB panoramas with room layout annotations provided by \cite{zou2018layoutnet}. The panoramic images are captured from large-scale indoor environments such as offices, classrooms, and corridors. This is a more challenging dataset since it has more occlusions on the floor boundaries and the images have a smaller vertical field of view.

\begin{table}[t]
\centering
 \begin{tabular}{||c c c c||} 
 \hline
 Method & Pixel & Corner & 3D IoU \\ [0.5ex] 
  &Error ($\%$)&Error ($\%$)&\\[0.5ex]
 \hline\hline
 PanoContext \cite{zhang2014panocontext} & 4.55 & 1.60 & 67.23 \\ 
 \hline
 CFL \cite{fernandez2020corners} & 2.49 & 0.79 & 78.79 \\
 \hline
 LayoutNet \cite{zou2018layoutnet} & 3.34 & 1.06 & 74.48 \\
 \hline
 DuLa-Net \cite{yang2019dula} & - & - & 77.42 \\
 \hline
 HorizonNet \cite{sun2019horizonnet} & 2.7 & 0.82 &  79.8\\
 \hline
 \textbf{Ours} &  \textbf{2.2}&  \textbf{0.75}&  \textbf{83.02}\\
 \hline
\end{tabular} \\~\\
 \label{tab:Table1}
\caption{Cuboid layout estimation evaluation on PanoContext Dataset \cite{zhang2014panocontext} (Training set - PanoContext \cite{zhang2014panocontext}).}
\end{table}

\begin{table}[t]
\centering
 \begin{tabular}{||c c c c||} 
 \hline
 Method & Pixel & Corner & 3D IoU \\ [0.5ex] 
  & Error ($\%$)& Error ($\%$)&\\[0.5ex]
 \hline\hline
 LayoutNet \cite{zou2018layoutnet} & 3.18 & 1.02 & 75.12 \\
 \hline
 HorizonNet \cite{sun2019horizonnet} &  2.6 & 0.79 & 80.2 \\
 \hline
 \textbf{Ours} & \textbf{2.10} & \textbf{0.69} & \textbf{84.5} \\  
 \hline
\end{tabular} \\~\\
\caption{Cuboid layout estimation evaluation on PanoContext Dataset \cite{zhang2014panocontext} (Training set - PanoContext \cite{zhang2014panocontext} + Stanford 2D-3D \cite{armeni2017joint}). }
\label{tab:Table2}
\end{table}

\subsection{Setup and Implementation Details}\label{subsec:training_Details}

We follow the same train/val/test split as LayoutNet \cite{zou2018layoutnet} for both the datasets PanoContext \cite{zhang2014panocontext} and Stanford 2D-3D \cite{armeni2017joint} and use the same experimental protocol described in \cite{sun2019horizonnet} for training the baseline method. Training and test set images are pre-processed by the panoramic image alignment method proposed in \cite{zou2018layoutnet}. PanoStretch data augmentation \cite{sun2019horizonnet} is used to augment the training data by stretching the panorama images along the axes in 3D space. The main idea of PanoStretch data augmentation \cite{sun2019horizonnet} is to convert the pixels of the equirectangular image to 3D space and multiply their X, Y, Z coordinates with separate hyperparameters (augmentation parameters). The stretched points can then be projected back to form the final image. Exact details can be found in  \cite{sun2019horizonnet}.

To predict the position of the floor-wall boundary ($y_f$) and ceiling-wall boundary ($y_c$) we use L1 loss for the learning, whereas for the prediction of the wall-wall boundary ($y_w$) we use binary cross-entropy loss. The optimizer used is Adam with a learning rate of 0.0003. We train all our networks by using the pretrained Imagenet weights for 150 epochs with a batch size of 4. It takes around 12 hours to finish the training on a single NVIDIA GTX 1080 GPU. \footnote{Training time can be further optimized by optimizing spherical CNN implementation on CUDA. The training time in the case of standard convolutions is 3x faster.}

\subsection{Experimental Results}\label{subsec:results}
We present the results on two benchmark datasets described in Sec. \ref{subsec:data} and compare our model with the current state-of-the-art models in Table. 1 - Table. 4.
We perform both quantitative (Sec. \ref{subsec:quantitative}) and qualitative evaluation (Sec. \ref{subsec:quantitative}). Finally in Sec.  \ref{subsec:ablation}, we conduct ablation studies to highlight the gain in performance due to the proposed sphere convolution formulation.

\begin{table}[t]
\centering
 \begin{tabular}{||c c c c||} 
 \hline
 Method & Pixel & Corner & 3D IoU \\ [0.5ex] 
  &Error ($\%$)&Error ($\%$)&\\[0.5ex]

 \hline\hline
 LayoutNet \cite{zou2018layoutnet} & 2.70 & 1.04 & 76.33 \\
  \hline
 DuLa-Net \cite{yang2019dula} & - & - & 79.63 \\
 \hline
 HorizonNet \cite{sun2019horizonnet} & 2.5 & 0.97 &  77.2\\ 
 \hline
 \textbf{Ours} & \textbf{2.37} & \textbf{0.78} & \textbf{81.2} \\ 
 \hline

\end{tabular} \\~\\
\caption{Cuboid layout estimation evaluation on Stanford 2D-3D Dataset \cite{armeni2017joint} (Training set - Stanford 3D-3D \cite{armeni2017joint}).}
\label{tab:Table3}
\end{table}

\begin{table}[t]
\centering
 \begin{tabular}{||c c c c||} 
 \hline
 Method & Pixel & Corner & 3D IoU \\ [0.5ex] 
  &Error ($\%$)&Error ($\%$)&\\[0.5ex]
 \hline\hline
 LayoutNet \cite{zou2018layoutnet} & 2.42 & 0.92 & 77.51 \\
 \hline
 HorizonNet \cite{sun2019horizonnet} & 2.36 & 0.77 & 80.8 \\ 
 \hline
 \textbf{Ours} & \textbf{2.14} & \textbf{0.68} &  \textbf{83.4}\\ 
 \hline
\end{tabular} \\~\\
\caption{Cuboid layout estimation evaluation on Stanford 2D-3D Dataset \cite{armeni2017joint} (Training set - PanoContext \cite{zhang2014panocontext} + Stanford 2D-3D \cite{armeni2017joint}).}
\label{tab:Table4}
\end{table}

\begin{figure*}[!t]
    \centering
    \includegraphics[width=\textwidth]{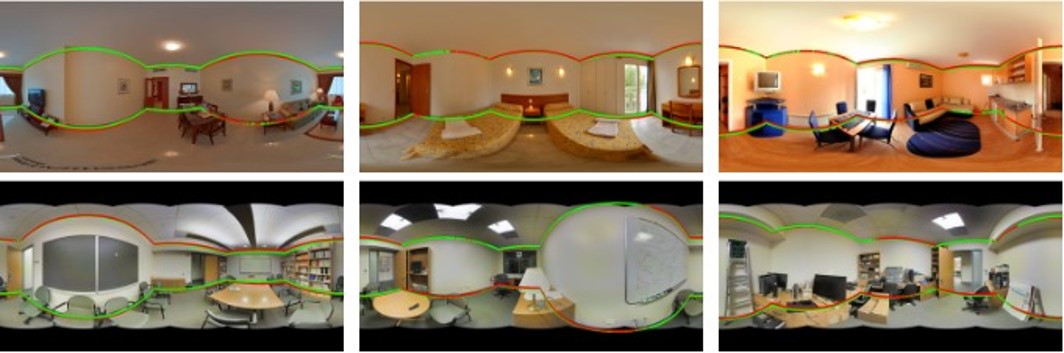}
    \caption{Qualitative results for room layout estimation on PanoContext \cite{zhang2014panocontext} (top) and Stanford 2D-3D \cite{armeni2017joint} (bottom). Each image was randomly sampled from the dataset. Our model's prediction is highlighted in \textcolor{red}{red} color whereas the ground truth is highlighted in \textcolor{green}{green} color. Best viewed in color.}
    \label{fig:stanford_Results}
\end{figure*}

\begin{figure*}[!t]
    \centering
    \includegraphics[width=\textwidth,height=5.5cm]{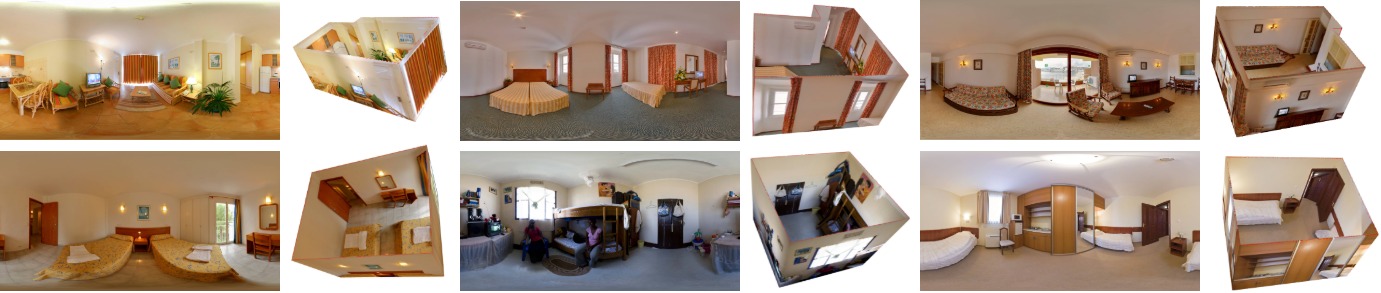}
    \caption{Qualitative results for the non-cuboid layout prediction (top row) and the cuboid layout prediction (bottom row). Best viewed in color.}
    \label{fig:layout}
\end{figure*}

\subsubsection{Quantitative Evaluation}\label{subsec:quantitative}

We measure the quantitative performance based on three standard metrics: 3D intersection over union between the predicted 3D reconstructed layout and ground truth, pixel error between predicted and ground truth surface class, and corner error which measures the euclidean distance between predicted and ground truth corners. Apart from the architectural difference, existing literature differ in the input resolution and augmentation techniques. Clara \textit{et al.} \cite{fernandez2020corners} and Yang \textit{et al.} \cite{yang2019dula} use input image of resolution $ 256 \times 512 $ whereas Zou \textit{et al.} \cite{zou2018layoutnet}, Sun \textit{et al.} \cite{sun2019horizonnet} and our method uses input resolution of $ 512 \times 1024$. Clara \textit{et al.} \cite{fernandez2020corners} is trained with random erase augmentation technique, while our method and Sun \textit{et al.} \cite{sun2019horizonnet} are both trained with the PanoStretch augmentation technique. Throughout our evaluation from Table 1 - Table 4, we compare with the state-of-the-art results that we were able to reproduce from the open-source codes. 

Based on the evaluation metrics pixel error ($\%$), corner error ($\%$), and 3D IoU we can see that our method is the new state-of-the-art approach and  outperforms all prior methods by $\approx$ 4$\%$ on both benchmarks PanoContext \cite{zhang2014panocontext} and Stanford 2D-3D \cite{armeni2017joint}. The comparison with more relevant approach \cite{fernandez2020corners} validates our hypothesis since they are only other work using spherical convolutions for layout estimation. However Clara \textit{et al.} \cite{fernandez2020corners} use spherical convolution with inverse gnomic projection and reports 3D IoU, which is $\approx$ ~5$\%$ lower compared to our method on both PanoContext \cite{zhang2014panocontext} and Stanford 2D-3D \cite{armeni2017joint} benchmarks. 

Although our method uses a different projection type for spherical convolution, in ablation studies we show our network's results with the inverse gnomic projection are similar to the projection type used in \cite{fernandez2020corners} despite this our network achieves  $\approx$ 4$\%$ better performance than Clara \textit{et al.} \cite{fernandez2020corners}. Therefore we hypothesis that the boost in our performance is due to incorporation of spherical convolution with better representation architecture. To validate the importance of spherical convolution alone, we perform several experiments as described in Sec. \ref{subsec:ablation}.

\subsubsection{Qualitative Evaluation}\label{subsec:qualitative}

We present the qualitative results in the form of room layout maps (Fig.\ref{fig:stanford_Results}) and 3D layouts of both non-cuboid and cuboid-shaped rooms (Fig. \ref{fig:layout}). The non-cuboid rooms in Stanford 2D-3D \cite{armeni2017joint} and PanoContext \cite{zhang2014panocontext} are labelled as cuboids, thus making it difficult for our model to learn non-cuboid rooms. To overcome this, we use the 65 general room layouts re-labeled by Sun \textit{et al.} \cite{sun2019horizonnet} in the train split to fine-tune our network. The samples in Fig. \ref{fig:layout} show that our network is capable of generating both non-cuboid (``L-shaped") room layouts as well as cuboid room layouts. 

From Fig. \ref{fig:stanford_Results} we can observe the obvious similarity between model's predictions and ground truth. One important aspect of our model is the capability to detect the corner while estimating the boundary line even when the corner is hidden (such as corner hidden behind a door in Fig. \ref{fig:stanford_Results}). We believe that our model representation combined with Bi-GRU to understand context over longer horizon, leads to better prediction of corners.


\begin{table}[t]
\centering
 \begin{tabular}{||c c c c||} 
 \hline
 Method & Pixel & Corner & 3D IoU \\[0.5ex] 
 &Error ($\%$)&Error ($\%$)&\\[0.5ex]
 \hline\hline
 Standard Conv & 2.6 &  0.79 &  81.4 \\ [1ex] \hline
 Sphere Conv   & 2.09  &  0.669 & 84.65 \\[1ex]   
 (Inv. gnomic proj.)& &  & \\[1ex]
 \hline
 \textbf{Ours} & \textbf{2.06} & \textbf{0.662} & \textbf{86.15} \\ [1ex] 
 \hline
\end{tabular} \\~\\
\caption{Comparison between standard convolution, spherical convolution (with inverse gnomic projection) and spherical convolution (with inverse equirectangular projection) on the PanoContext \cite{zhang2014panocontext} + Stanford 2D-3D dataset \cite{armeni2017joint}}
\label{tab:Table6}
\end{table}

\begin{table}[htb!]
\centering
 \begin{tabular}{||c c c c||} 
 \hline
 Method & Pixel & Corner & 3D IoU \\[0.5ex] 
 &Error ($\%$)&Error ($\%$)&\\[0.5ex]
 \hline\hline
 Standard Conv & 2.56 & 0.79 & 81.4 \\ [0.5ex] 
  \footnotesize (ResNet34) & &  & \\ \hline
 \textbf{Spherical Conv} & \textbf{2.2} & \textbf{0.67} &  \textbf{85.7}\\ [0.5ex] 
   \footnotesize \textbf{(ResNet34)} & &  & \\ \hline
 \hline
  Standard Conv & 2.53 & 0.77 & 82.1  \\ [0.5ex]
    \footnotesize (ResNet101) & &  & \\ \hline
 \textbf{Spherical Conv }& \textbf{2.05} & \textbf{0.65 }& \textbf{86.3} \\ [0.5ex]
   \footnotesize \textbf{(ResNet101) }& &  & \\ \hline
 \hline
  Standard Conv & 2.52 & 0.76 & 82.4 \\ [0.5ex]
    \footnotesize (ResNet151) & &  & \\ \hline
 \textbf{Spherical Conv} & \textbf{2.04} & \textbf{0.65} & \textbf{86.5} \\ [0.5ex]
   \footnotesize \textbf{(ResNet151)} & &  & \\ \hline

 \hline
\end{tabular} \\~\\
\caption{Comparison between standard convolution and proposed spherical convolution for 3 different networks of the ResNet family. Evaluation done on both PanoContext \cite{zhang2014panocontext} and Stanford 2D-3D \cite{armeni2017joint}.}
\label{tab:Table7}
\end{table}
\begin{figure}
    \centering
    \includegraphics[width=\linewidth]{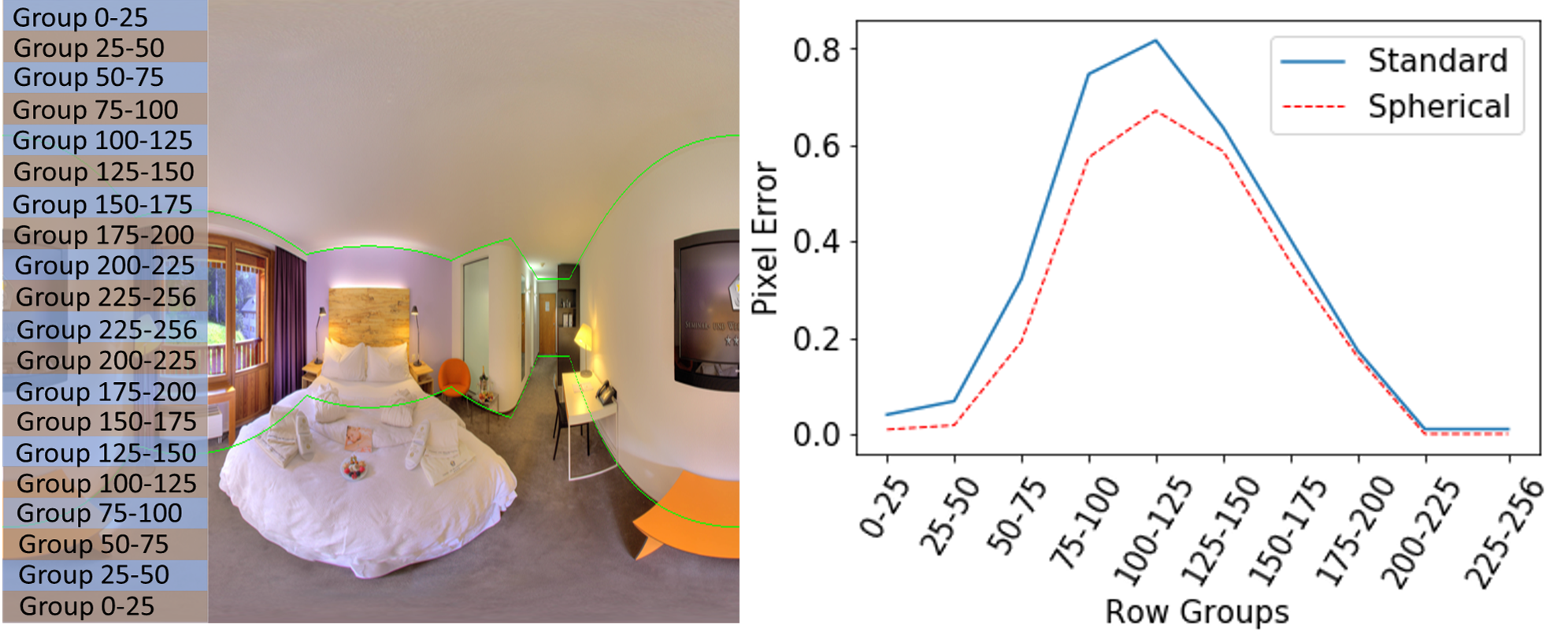}
    \caption{Input images are divided into row groups (left image shows a sample for reference) based on distance from the poles. We calculate pixel error in these row groups to show the difference in performance across different regions of the image. Results evaluated on PanoContext \cite{zhang2014panocontext} and Stanford 2D-3D  \cite{armeni2017joint}.}
    \label{fig:line_chart}
\end{figure}

\subsection{Ablation Study}\label{subsec:ablation}
In Table. 5, we compare the results with standard convolution, spherical convolution with inverse gnomic projection \cite{coors2018spherenet} and our proposed spherical convolution with inverse equirectangular projection. It is evident from our results that both the spherical convolutions (inverse gnomic and inverse equirectangular projection) are better when compared with standard convolution. Thus validating the hypothesis that spherical convolutions are well suited for this problem and can efficiently handle the distortions in the equirectangular images. Since the property of equirectangular images are more inclined towards cylindrical projections rather than projections over tangent plane, inverse equirectangular projection offers rich representation and leads to improved performance ($\approx 2$ $\%$) than inverse gnomic projection (Table. 5).

While dealing with equirectangular images metric such as pixel error offer least information and fail to capture significance of the model, since error is calculated over entire dataset. It is important to know the various regions (i.e simple or complex regions) in image which lead to improvement or degradation in model performance. To incorporate this scenario we propose a new metric, which identifies the region or groups where spherical convolution performs better than standard convolution. In Fig. \ref{fig:line_chart}, we plot the pixel error ($\%$) observed in the test set for standard convolution and spherical convolution against row groups. The panoramic images are divided into different row groups based on distance from the poles (See Fig. \ref{fig:line_chart}), where each row group has a width of 25 rows.
As hypothesised the difference in the pixel space is highest ($\approx$25$\%$) when we are closer to the poles and the image regions where the ceiling-wall and floor-wall boundaries are likely to appear in majority of samples. The difference gradually decreases as we go towards the equator of the image (i.e simple region). This confirms to our assumption that majority of the distortion that our method removes are near the poles of the image, which we categorize as the difficult or complex regions for standard convolutions.

Finally, we input our proposed sphere convolution to the following networks of the ResNet family: ResNet-34 \cite{he2016deep}, ResNet-101 \cite{he2016deep}, and ResNet-151 \cite{he2016deep} and report the results in Table 6. It is evident that our approach is not restricted to any architecture and can improve performance across any convolution architecture. The proposed method is independent of model parameters and depth of the network, hence for complex tasks can also be extended to work with very deep networks.

\section{Conclusions}\label{sec:con}

We proposed a novel state-of-the-art approach which reduces the distortions in equirectangular images for the task of 360$\degree$ room layout recovery. In our knowledge this is the first work in room layout estimation that uses the equirectangular projection function to reduce the distortions. The proposed method, OmniLayout is computationally efficient and can also recover both cuboid shaped layouts as well non-cuboid shaped layouts (``L-shaped''). The experimental analysis and ablation study shows that OmniLayout significantly improves the performance on two room layout benchmark datasets, especially in the distortion heavy regions of the input panoramic images. 
{\small
\bibliographystyle{ieee_fullname}
\bibliography{zmain_arxiv}
}

\end{document}